\documentclass[runningheads]{llncs}
\usepackage[T1]{fontenc}
\usepackage{graphicx}
\newcommand{\mypar}[1]{\smallskip\noindent\textbf{#1.}}
\newcommand{\mypartwo}[1]{\vspace{0.5pt}\noindent\textit{#1.}}

\usepackage{rotating}
\usepackage{varwidth}
\usepackage{float}
\usepackage{needspace}

\usepackage[hidelinks]{hyperref}
\usepackage{booktabs}
\usepackage{multirow}
\usepackage{subcaption}
\usepackage{cite}
\usepackage{algorithm}
\usepackage{algorithmic}
\usepackage{enumitem}
\newcommand{\COMMENTTRIANGLE}{$\quad \triangleright$\ }

\usepackage{color}

\begin{document}
%

\title{A Divide-and-Conquer Approach for Modeling Arrival Times in Business Process Simulation}

\titlerunning{A Divide-and-Conquer Approach for Modeling Arrival Times}


%

\author{Lukas Kirchdorfer\thanks{Equal contribution}\inst{1,2} \and
Konrad Özdemir$^\star$\inst{2} \and \\
Stjepan Kusenic\inst{2}\and
Han van der Aa\inst{3}\and
Heiner Stuckenschmidt\inst{2}
}
\authorrunning{L. Kirchdorfer and K. Özdemir et al.}
%
\institute{SAP Signavio, Walldorf, Germany\\
\email{lukas.kirchdorfer@sap.com} \and
Data and Web Science Group, University of Mannheim, Germany\\
\email{\{konrad.oezdemir, heiner.stuckenschmidt\}@uni-mannheim.de} \and
Faculty of Computer Science, University of Vienna, Austria\\
\email{han.van.der.aa@univie.ac.at}}

\maketitle             

\begin{abstract}
Business Process Simulation (BPS) is a critical tool for analyzing and improving organizational processes by estimating the impact of process changes. A key component of BPS is the case-arrival model, which determines the pattern of new case entries into a process. Although accurate case-arrival modeling is essential for reliable simulations---as it influences waiting and overall cycle times---existing approaches often rely on oversimplified static distributions of inter-arrival times. These approaches fail to capture the dynamic and temporal complexities inherent in organizational environments, leading to less accurate and reliable outcomes.
To address this limitation, we propose \emph{Auto Time Kernel Density Estimation} (AT-KDE), a divide-and-conquer approach that models arrival times of processes by incorporating global dynamics, day-of-week variations, and intraday distributional changes, ensuring both precision and scalability. Experiments conducted across 20 diverse processes demonstrate that AT-KDE is far more accurate and robust than existing approaches while maintaining sensible execution time efficiency.

\end{abstract}


\section{Introduction}
Business Process Simulation (BPS) is a key tool to support the redesign of organizational processes and their underlying information systems~\cite{Aalst15}. 
By being able to establish \textit{digital process twins}~\cite{Dumas21}, BPS can be used to estimate the impact of process improvement ideas on performance indicators such as cycle time. This may include the implementation of a new system or the roll-out of a new process version. By providing such estimates in advance, process simulation has the potential to drastically improve the efficiency and reduce the risks of redesign efforts for decision-makers~\cite{FundamentalsOfBPM}. However, the effectiveness of BPS heavily relies on the availability of a simulation model that accurately mimics the real-world behavior of a process, since only accurate models can provide trustworthy insights into the impact of a redesign. Given that the manual construction of simulation models is time-consuming and error-prone \cite{Aalst15}, various automated approaches have been developed for discovering BPS models from historical execution data contained in event logs~\cite{RozinatMSA09,Camargo_2020,meneghello_RIMS,camargo_DSIM,Kirchdorfer2024}.
Usually, such data-driven simulators discover a process model and augment it with simulation parameters, such as a case-arrival model and activity processing times. 

In this work, we argue that existing data-driven BPS approaches have a clear limitation when it comes to the manner in which they model the arrival of new cases, a factor shown to critically impact simulation outcomes~\cite{MartinDC15}. Specifically, most approaches in BPS use a static distribution to represent the case-arrival model. However, static models cannot account for dynamic organizational environments with varying arrival rates over time, as their parameters are inherently time-invariant. Consequently, these models are unable to adapt to environmental changes driven by global seasonalities, local weekday patterns, or intraday fluctuations.
For instance, patient arrivals in a hospital emergency department vary considerably, spiking during flu season, decreasing on weekends, and surging in the mornings compared to late nights.
Failing to capture such variations in a case-arrival model means that simulation results obtained from it will not reflect realistic loads of the system, despite this having a tremendous impact on factors such as resource utilization, waiting time, and cycle time.

Therefore, we use this work to propose \emph{Auto Time Kernel Density Estimation} (AT-KDE), an approach for modeling arrival times in BPS. 
AT-KDE combines a divide-and-conquer strategy with Kernel Density Estimation (KDE)~\cite{kde} to iteratively segment the arrival time series and learn the segments' distributions non-parametrically. 
Thereby, it captures varying patterns at multiple time-related levels. Our approach enables precise and scalable arrival time modeling by explicitly accounting for global dynamics like seasonalities or drifts, day-of-week variations, and time-specific distributions within a day, most of which are ignored by existing approaches. 
Our extensive evaluation across 20 different processes demonstrates that AT-KDE is considerably more accurate and robust than existing approaches used in BPS and approaches from other domains.

The remainder starts by discussing related work in~\autoref{sec:rel_work}, before motivating our approach in~\autoref{sec:motivation}. Afterwards,~\autoref{sec:method} describes our approach for modeling and simulating arrival times with an experimental evaluation in~\autoref{sec:evaluation}. Finally,~\autoref{sec:conclusion} concludes our work.

\section{Related Work} \label{sec:rel_work} 

This section reviews work on modeling arrivals in BPS and in other domains.


\mypar{Approaches Applied in BPS}
Arrival modeling in BPS, so far, predominantly employs static approaches, with only one dynamic alternative.

\mypartwo{Static Approaches} 
Case arrivals are typically simulated by sampling from a fitted probability distribution of inter-arrival times and cumulatively adding those samples to attain the concrete case's arrival timestamps. 
The respective inter-arrival times are typically modeled using an exponential distribution with a constant rate~\cite{Aalst15}, as seen in Rozinat et al.'s simulation model~\cite{RozinatMSA09}. Martin et al.~\cite{MartinDC15} similarly adopt the gamma distribution---a generalization of the exponential. A more recent approach by Camargo et al. \cite{Camargo_2020} relaxes the apriori selection by allowing for a discovery of the best-fitting distribution from a predefined set of distributions. 
This has been adopted also by other BPS approaches~\cite{Kirchdorfer2024,Lopez-PintadoD22}.

\mypartwo{Dynamic Approaches}
Differently to fitting a distribution on inter-arrival times, 
two BPS approaches~\cite{camargo_DSIM,meneghello_RIMS} use \emph{Prophet}~\cite{TaylorL17} to decompose time series into trend, seasonality, holidays, and error components. It is used to predict hourly arrival counts, which are then uniformly distributed within each hour to form timestamps. Although it improves upon static approaches, \emph{Prophet} has notable limitations for process simulation, as we will discuss in \autoref{sec:motivation}.

\mypar{Approaches Used Beyond BPS} 
The simulation of process arrivals can also be viewed as a (dynamic) time series forecasting problem, relevant in many scenarios beyond BPS, with approaches categorized into \textit{Classical} and \textit{ML Models}.

\mypartwo{Classical Models}
Classical time series models provide a reliable basis for forecasting temporal events. A notable example is the \emph{Autoregressive Integrated Moving Average} (ARIMA) model, which predicts future values by leveraging past observations and error terms. Its robust framework is well established, though it requires careful stationarity checks and parameter tuning, limiting its out-of-the-box use~\cite{hamilton_time_series}. Despite this, ARIMA remains versatile and efficient; e.g., with applications in healthcare, finance, and energy~\cite{arima_survey}. Another established approach is the \emph{Hawkes Process}.
This nonhomogeneous, self-exciting point process (NPP) captures temporal clustering, making it ideal for scenarios where current events trigger future arrivals. It has gained significant attention in queuing systems~\cite{hawkes_in_queues}. However, due to its inherently parametric nature, we consider it to be \textit{static}. 

\mypartwo{ML Models}
In recent years, machine learning models have emerged as a powerful complement to classical approaches by directly modeling complex relationships in temporal data. The \emph{Long Short-Term Memory} (LSTM) network, a recurrent neural network variant, is widely used to capture long-term dependencies and forecast events in environments with complex temporal dynamics~\cite{lstm_survey}. Ensemble methods such as XGBoost have also been applied effectively to arrival time forecasting, offering competitive performance by exploiting non-linear interactions in the data~\cite{PortoF24}. More recently, frameworks like Chronos, which build upon advances in foundation time series modeling, have been proposed to leverage large-scale data and pre-trained representations~\cite{Ansari2024}. 

Our experiments will show that static methods ignore pattern variability and dynamic methods lack robustness and incur high computational costs; in contrast, AT-KDE captures process dynamics both efficiently and robustly.





\section{Motivation}
\label{sec:motivation}
In this section, we highlight the critical impact of case arrivals on the accuracy of BPS outcomes. Using a loan application process as an example, we examine how different types of dynamics influence arrival patterns and complicate the extraction of a reliable arrival model from process data. We then show that existing arrival modeling approaches used in BPS fail to account for these dynamics, leading to inaccurate arrival estimates and, thus, poor simulation accuracy.

\begin{figure}[htbp]
    \centering
    \includegraphics[trim = 0mm 0mm 0mm 0mm, clip, width=\columnwidth]{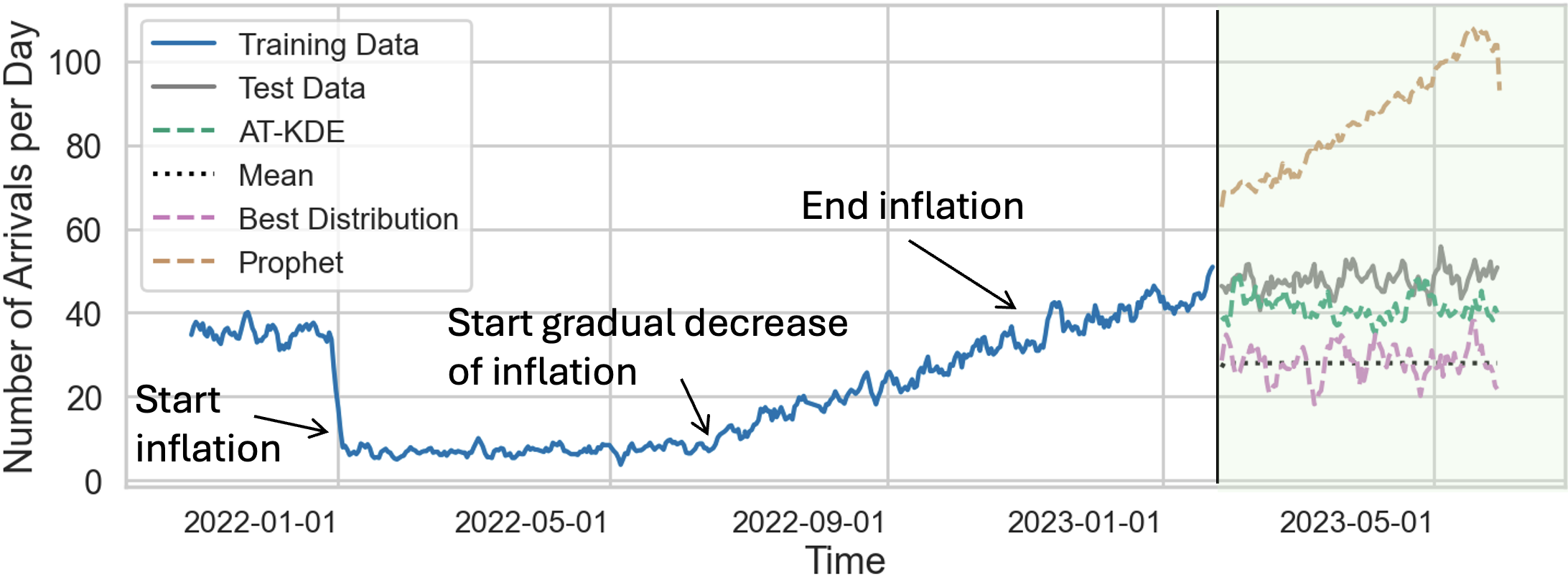}
    \caption{7-Day rolling average of arrival count in the loan application process.}
    \label{fig:corona}
\end{figure}

\mypar{Understanding Arrival Behavior}
To illustrate the intricate complexities within arrival data, we consider a loan application process (cf.~\cite[Chapter 10.8]{FundamentalsOfBPM}) during a period of changing interest rates (data available in \cite{atkde_data_repo_zenodo})
and focus on arrivals of new applications, visualized in \autoref{fig:corona}. Within this scenario, we can expect various types of dynamics leading to variations in arrival rates over time: 
\begin{itemize}[topsep=0pt]
    \item \textit{Intraday Dynamics.} The arrival pattern can change over the course of a single day. For instance, customers may be more likely to submit loan applications in the morning, with fewer arrivals in the evening or overnight.

    \item \textit{Weekday Dynamics.} The arrival pattern can vary across the days of the week. For instance, loan applications may peak on Mondays as customers submit documents prepared over the weekend, decline gradually during the week, and drop on weekends---especially if applications require in-person visits.

    \item \textit{Global Dynamics.} Broader trends and external factors can alter arrival rates over extended periods. In our scenario, rising interest rates driven by inflation due to a market crash may lead to a sudden decline in loan applications, followed by a gradual recovery as economic conditions stabilize.

\end{itemize}

\noindent Note that we define global dynamics to include all variations beyond weekly patterns, such as seasonal trends (e.g., peaks at the beginning of the month), long-term drifts (e.g., evolving financial policies), and one-off events (e.g., economic shocks). While some of these changes are predictable, others are sudden and disruptive. By explicitly modeling even one-off events, we separate their impact from recurring patterns, leading to more accurate future arrival estimates without distortions by past anomalies.

\mypar{Limitations of Existing Approaches in BPS}
Given the above illustration, we postulate that an effective approach for deriving an arrival model from data should take these three dynamics into account, i.e., it needs to be able to consider potentially varying arrival patterns on the intraday, weekday, and global level.

Moreover, from this perspective, we argue that existing approaches used in BPS fail to do so, leading to poor arrival estimates as shown in the green-shaded area of \autoref{fig:corona}. Static approaches ignore all three dynamics by assuming a single distribution over inter-arrival times. For instance, \emph{Best Distribution}\cite{Camargo_2020}, which fits a parameterized distribution to the inter-arrival times just fluctuates around the naive \emph{Mean} baseline. \emph{Prophet}, a dynamic approach, appears to capture weekday and global variations to some extent but neglects intraday patterns, as its common implementation in BPS approaches \cite{camargo_DSIM,meneghello_RIMS} estimates arrivals per time unit and distributes them randomly, lacking a direct mechanism for precise timestamp estimation. In our loan process, \emph{Prophet} greatly exaggerates the gradual trend, which we suspect to be a general problem due to a lack of autoregressive components~\cite{TaylorL17}. In contrast, the green line representing our AT-KDE approach (detailed in the next section) shows that more accurate arrivals can be generated by appropriately taking dynamics at different levels into account.


\mypar{Impact on BPS Accuracy} 
We compare arrival modeling approaches in the loan application process to demonstrate how poor arrival estimates—e.g., those that ignore temporal dynamics—can degrade simulation accuracy. We employ \emph{AgentSimulator}\cite{Kirchdorfer2024} to discover a BPS model and simulate process executions, evaluating results against a held-out test set using the current state-of-the-art BPS evaluation framework~\cite{chapela2023}. Focusing on time and congestion perspectives, we report distribution distances for cycle time (CTDD), absolute event time (AEDD), circadian event time (CEDD), and relative event time (REDD).

\vspace{-2em}
\begin{table}[htbp]
\centering
\setlength\tabcolsep{6pt} 
\caption{Simulation accuracy for different arrival models (lower is better).}
\label{tab:loan_app}
\begin{tabular}{l|rrrr}
\toprule
 Arrival approach & CTDD & AEDD & CEDD & REDD  \\
\midrule
Mean & 114.53 & 146.86 & 0.20 & 118.23 \\
Best Distribution & 112.67 & 145.47 & 0.24 & 115.97 \\
Prophet & 504.65 & 605.26 & 0.83 & 433.67 \\
AT-KDE & \textbf{54.08} & \textbf{85.59} & \textbf{0.12} & \textbf{56.18} \\

\bottomrule
\end{tabular}
\end{table}


\vspace{-1em}

\autoref{tab:loan_app} suggests that (i) simulation quality varies greatly depending on the chosen arrival approach, and (ii) AT-KDE consistently achieves the most accurate results. For instance, AT-KDE yields a CTDD of 54.08, caused by an average simulated cycle time of 119 hours, compared to 143 hours in the ground truth, 29 hours with \emph{Best Distribution}, and 640 hours with \emph{Prophet}. These deviations reflect arrival estimation errors (cf. \autoref{fig:corona}): \emph{Best Distribution} underestimates arrivals, shortening cycle times, while \emph{Prophet} overestimates them, inflating delays.
Now, consider a simulation used to assess whether hiring an additional employee would reduce application-to-decision times.
If the arrival model significantly misrepresents the actual workload---as with \emph{Best Distribution} or \emph{Prophet}---the simulation's outcome may be misleading, potentially causing costly or ineffective staffing decisions. These results highlight the critical need for accurate arrival modeling to ensure simulation outcomes are realistic. 

\section{Approach}
\label{sec:method}
This section presents our AT-KDE approach for modeling and simulating arrival times in BPS. Its input, output, and main steps are as follows:

\mypar{Input} 
Our approach takes as input an \textit{arrival dataset}, denoted by $\mathcal{D} = (\mathbf{t}^i)_{i=1}^N$. This dataset is a sequence of $N$ sequences,  where each $\mathbf{t}^i$ represents the sequence of arrival timestamps on a day $i \in \{1,\dots,N\}$. Note that the length of $\mathbf{t}^i$ varies.

\mypar{Output} Similar to its input, our approach generates as output another sequence of sequences of arrival timestamps $\mathcal{D}' = (\Tilde{\mathbf{t}}^i)_{i=1}^{\Tilde{N}}$ over $\Tilde{N}$ days.

\begin{figure}[t!]
    \centering
    \includegraphics[trim = 0mm 320mm 0mm 10mm, clip, width=\columnwidth]{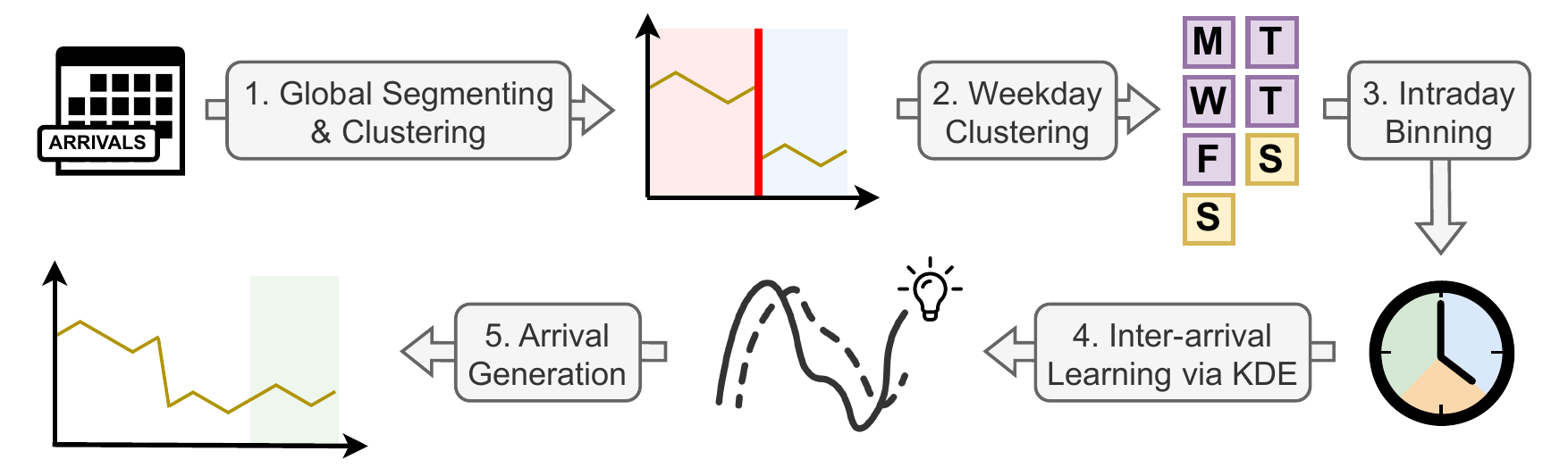}
    \caption{Workflow of the modeling methodology of our AT-KDE approach.}
    \label{fig:flowchart}
\end{figure}

\mypar{Approach Steps} The overarching goal is to simulate realistic arrival times by discovering a case-arrival model that explicitly captures the dynamic nature of organizational environments. To achieve this, our \emph{divide-and-conquer} AT-KDE approach consists of five main steps, as visualized in \autoref{fig:flowchart}. 
The first three steps make up the \emph{divide} phase, where the arrival dataset is partitioned into subsets to address the different types of dynamics outlined in \autoref{sec:motivation}. In the \emph{conquer} phase, steps 4 and 5 focus on learning a separate KDE model for each subset and combining these models to generate new arrivals.
\begin{enumerate}[noitemsep,topsep=0pt]
    \item \textit{Global Segmenting and Clustering.}
    Given the arrival dataset $\mathcal{D}$, we account for \textit{global dynamics} by identifying change points that signal potential drifts or seasonalities in the data. These change points divide the dataset into segments, capturing periods of consistent arrival patterns. To group similar patterns, the segments are clustered into a set of $J\geq 1$ segment clusters $\{D_1, \dots, D_J\}$, where each cluster contains one or more segments from $\mathcal{D}$.
    \item \textit{Weekday Clustering.} 
    Given a segment cluster $D_j$, we account for \textit{weekday dynamics} by grouping weekdays with similar arrival patterns. Thus, we divide the global cluster $D_j$ into weekday clusters $\{W_1^j, \dots, W_{K_j}^j\}$. If no clusters are found we obtain the upper bound (number of days in a week): $K_j = 7$. 
    
    \item \textit{Intraday Binning.} 
    Given a weekday cluster $W_k^j$, we account for \textit{intraday dynamics} by dividing a day $\mathbf{t}^i\in W_k^j$ into $L\in \mathbb{N}$ equally long bins.
    
    \item \textit{Inter-arrival Learning via KDE.} 
    The arrival dataset $\mathcal{D}$ is now divided into multiple disjoint subsets. Each of those contains sequences of arrival timestamps that exhibit similar characteristics. Next, for each subset, a separate KDE-model is fitted to the inter-arrival times, resulting in an ensemble of KDE-models that collectively capture the diverse dynamics of the data.
    
    \item \textit{Arrival Generation.} 
    Finally, we iteratively sample from the  KDE-model ensemble to generate new arrival timestamps for the simulation period, resulting in the output arrival dataset $\mathcal{D'}$.
\end{enumerate}

\subsection{Step 1: Global Segmenting and Clustering}
\label{sec:global_property}
The first step focuses on identifying global dynamics in the arrival dataset $\mathcal{D}$, such as seasonalities or concept drifts. As outlined in Algorithm~\ref{alg:segmentation}, this is achieved by three key phases: \emph{Data Transformation} transforms the arrival dataset into a higher-level representation, enabling the detection of meaningful change points and clustering of the resulting segments based on statistical similarity in the \emph{Segmentation and Clustering} phase. Finally, the \emph{Solution Validation} phase evaluates the admissibility of the identified segment clusters by checking if they meet a set of predefined requirements. 
\begin{algorithm}[htbp]
\caption{ClusterGlobalSegments}
\label{alg:segmentation}
\begin{algorithmic}[1]
    \REQUIRE Arrival dataset $\mathcal{D}$, window size $\omega$, max clusters $k_{\max}$, sensitivity range $Z$
    \ENSURE Set of global segment clusters $\mathcal{G} = \{D_1, \dots, D_J\}$, cluster labels $\mathcal{L}$
    \STATE \COMMENTTRIANGLE{\textbf{Data Transformation}}
    \STATE $M \gets \text{GetArrivalCountSequence}(\mathcal{D})$ \label{line:transform}
    \STATE $M_{A} \gets \text{ComputeMovingAverages}(M, \omega)$ \label{line:moving_avg}
    \STATE $\Lambda \gets \text{ComputeSlidingWindowDifferences}(M_{A}, \omega)$ \label{line:sliding_windows}
    
    \STATE \COMMENTTRIANGLE{\textbf{Segmentation and Clustering}}
    \FOR{$z \in Z$} \label{line:loop}
        \STATE $\mathcal{C} \gets \text{DetectChangePoints}(\Lambda, z)$ \label{line:outliers}
        \STATE $\mathcal{S} \gets \text{GetSegments}(\mathcal{C}, \mathcal{D})$ \label{line:segments}
        \STATE $\mathcal{G}, \mathcal{L} \gets \text{ClusterSegments}(\mathcal{S})$ \label{line:clustering}
        \STATE $len_s \gets \text{GetSegmentLengthsInDays}(\mathcal{S})$
        \STATE \COMMENTTRIANGLE{\textbf{Solution Validation}}
        \IF{$\min(len_s) < \omega$ \textbf{or} $|\mathcal{S}| < 2$ \textbf{or} $|\text{unique}(\mathcal{L})| \geq k_{\max}$}
            \IF{$z == \text{FinalElementOf}(Z)$}
                \STATE $\mathcal{G} \gets \{\mathcal{D}\}; \mathcal{L} \gets \langle1\rangle$ \label{line:no_seg}
            \ENDIF
        \ELSE
            \STATE \textbf{break}
        \ENDIF
    \ENDFOR
    \RETURN $\mathcal{G}, \mathcal{L}$ \label{line:return2}
\end{algorithmic}
\end{algorithm}

\mypar{Data Transformation}
First, we aggregate the arrival dataset $\mathcal{D}$ into a sequence of daily arrival counts $M$, where each $M_i = |\mathbf{t}^i|$ represents the number of arrivals on day $i$, for $i = 1, \dots, N$ (Line~\ref{line:transform}). This aggregation forms the basis for detecting global dynamics.
Next, we compute the moving averages $M_{A}$ based on $M$ using a window size $\omega\in\mathbb{N}$ (Line~\ref{line:moving_avg}). This smoothing reduces noise and highlights trends in the arrival counts. The moving average at position $i$ is calculated as $M_{A,i} = \omega^{-1} \sum_{k=i}^{i+\omega-1} M_k$, $i = 1, \dots, N - \omega + 1$. 
We then calculate the sequence of sliding window differences $\Lambda$ based on $M_{A}$ (Line~\ref{line:sliding_windows}), which measures the average change between two consecutive moving windows: $\Lambda_{i} = M_{A,i+\omega} - M_{A,i},$ for $i = 1, \dots, N - 2\omega + 1$. This sequence captures the rate of change in $M_{A}$, where larger absolute values indicate more significant shifts in arrival patterns.

\mypar{Segmentation and Clustering}
Given the sequence of sliding window differences $\Lambda$, we proceed with identifying change points and clustering the resulting segments. The segment clustering serves two main purposes: it mitigates variance inflation due to data scarcity by combining data from similar segments, and it reveals global patterns given by the ordering of segment clusters, which is essential for simulating new arrival data $\mathcal{D}'$ in later steps.

\textit{DetectChangePoints}: We detect change points in the manner of outlier detection: We begin with some sensitivity parameter value $z \in Z\subseteq (0,1]$, controlling the strictness of the outlier detection, and compute a change factor $\text{CF} = 1.5 \times \text{IQR} \times z$, where $\text{IQR} = Q_3 - Q_1$ is the interquartile range given by the first ($Q_1$) and third ($Q_3$) quartiles of $\Lambda$. The factor of $1.5$ is a standard rule-of-thumb from box plot construction that defines the whiskers outside of the IQR to flag potential outliers, balancing sensitivity to extremes with robustness.
Then, indices where $\Lambda_{i}$ falls outside the range $R = [Q_1 - \text{CF}, Q_3 + \text{CF}]$ are identified as candidate change points: $\mathcal{\Tilde{C}} = \{ i \mid \Lambda_{i} \notin R\}$.
To assign only one change point to each drift, we constrain $\mathcal{\Tilde{C}}$: For each set of consecutive outlier-indices $G \subseteq \tilde{\mathcal{C}}$, we select the index $i \in G$ with the maximum absolute value of $\Lambda_{i}$. The set of change points is then defined as $\mathcal{C} = \{ i \in G \mid i = \arg\max_{j \in G} |\Lambda_{j}| \}$.

\textit{GetSegments}: The identified change points $\mathcal{C}$ partition $\mathcal{D}$ into a collection of disjoint segments $\mathcal{S} = \{\tilde{D}_1, \tilde{D}_2, \dots\}$ (Line~\ref{line:segments}). Each $\tilde{D}_s$ corresponds to a period between two change points and reflects a consistent arrival pattern. For instance, in our motivating loan application process in \autoref{fig:corona}, three change points are detected during the training period: \emph{Start inflation, Start gradual decrease of inflation}, and \emph{End inflation}, resulting in four adjacent segments.

\textit{ClusterSegments}: We next group segments with similar arrival patterns through clustering (Line~\ref{line:clustering}).
To do this, we characterize each segment via six statistical features: the average number of arrivals per day, the 25th and 75th percentiles of daily arrivals, the standard deviation of inter-arrival times, and the 25th and 75th percentiles of inter-arrival times. After standardizing these features (removing the mean and scaling to unit variance), and to avoid assuming the number of clusters apriori, we apply DBSCAN~\cite{dbscan} to label segments. Each unique label in $\mathcal{L}$ corresponds to a cluster $D_j \in \mathcal{G}$, $j \in \{1,\dots,J\}$, $J \leq |\mathcal{S}|$, denoting the disjoint union of all segments whose arrival sequences exhibit similar patterns. 

\mypar{Solution Validation}
Finally, we assess the admissibility of the segmentation and clustering results (Lines~10--17). Each segment must span at least $\omega$ days to capture meaningful global dynamics---shorter patterns are addressed in subsequent steps---ensuring segments are of sufficient length. 
Additionally, there must be at least two segments ($|\mathcal{S}|\geq 2$) to identify data changes, and fewer than $k_{\max}$ clusters to ensure each provides enough data for robust model fitting following subsequent steps. If these conditions are not met for the current sensitivity parameter $z$, we proceed to the next value in $Z$. If no value in $Z$ yields an admissible solution, we conclude that meaningful global change points are absent and set $\mathcal{G} = \{\mathcal{D}\}$ with a single cluster label $\mathcal{L} = \langle 1 \rangle$ (Line~\ref{line:no_seg}).

As a result, this step divides the entire dataset into statistically more coherent and disjoint subsets: $\mathcal{D} = \biguplus_{j=1}^J D_j$. For example, in our loan application process (cf. \autoref{fig:corona}), the segments [\textit{Begin-of-data}, \textit{Start inflation}] and [\textit{End inflation}, 
 \textit{End-of-data}] fall into the same cluster based on similar global dynamics.

\subsection{Step 2: Weekday Clustering} \label{sec:weekday_clustering}
Step 2 addresses \emph{weekday dynamics} within each global segment cluster identified in the first step. Specifically, for each global cluster $D_j$, we group weekdays with similar arrival patterns into weekday clusters $\{W_1^j, \dots, W_{K_j}^j\}$, where $K_j \leq 7$, $j =1, \dots, J$. This allows us to model day-of-week variations effectively. Algorithm~\ref{alg:weekday_clustering} outlines this process, which is divided into two phases: 

\begin{algorithm}[!htbp]
\caption{ClusterWeekdays}
\label{alg:weekday_clustering}
\begin{algorithmic}[1]
    \REQUIRE Global clusters $\mathcal{G} = \{D_1, \dots, D_J\}$
    \ENSURE Set of weekday clusters $\mathcal{W}$
    \FOR{each global cluster $D_j$} \label{line:global_cluster_loop}
        \STATE \COMMENTTRIANGLE{\textbf{Weekday Data Extraction}}
        \FOR{each weekday $w \in \{1, 2, \dots, 7\}$}
            \STATE $\mathcal{T}_w^j \gets \{ \mathbf{t}^i \in D_j \mid \text{weekday}(\mathbf{t}^i) = w \}$ \label{line:weekday_extraction}
            \STATE \COMMENTTRIANGLE{\textbf{Feature Computation and Clustering}}
            \IF{$\mathcal{T}_w^j \neq \emptyset$}
                \STATE $\mathbf{f}_w^j \gets \text{ComputeStatistics}(\mathcal{T}_w^j)$ \label{line:compute_statistics}
            \ENDIF
        \ENDFOR
        \STATE $\mathbf{F}^j \gets \text{StandardizeFeatures}(\{ \mathbf{f}_w^j \mid \mathcal{T}_w^j \neq \emptyset \})$ \label{line:standardize_features}
        \STATE $\text{WTC}_j \gets \text{WardClustering}(\mathbf{F}^j)$ \COMMENTTRIANGLE{Mapping from weekdays to cluster labels} \label{line:ward_clustering}
        \FOR{each cluster label $k$ in $\text{WTC}_j$}
            \STATE $W_k^j \gets \{ \mathbf{t}^i \in D_j \mid \text{WTC}_j(\text{weekday}(\mathbf{t}^i)) = k \}$ \label{line:weekday_cluster_definition}
        \ENDFOR
        \FOR{each weekday $w \in \{1, 2, \dots, 7\}$}
            \IF{$\mathcal{T}_w^j = \emptyset$}
                \STATE Assign $w$ to the special cluster $W_{\text{NoData}}^j$ \label{line:nodata_assignment}
            \ENDIF
        \ENDFOR
    \ENDFOR
    \RETURN $\mathcal{W} = \bigcup_{j=1}^{J} \{W_1^j, \dots, W_{K_j}^j\}$
\end{algorithmic}
\end{algorithm}

\mypar{Weekday Data Extraction}
For each weekday $w \in \{1, 2, \dots, 7\}$, we extract all arrival times $\mathbf{t}^i$ that occur on that weekday within the global cluster $D_j$ (Line~\ref{line:weekday_extraction}): $\mathcal{T}_w^j = \{ \mathbf{t}^i \mid \mathbf{t}^i \in D_j,\ \text{weekday}(\mathbf{t}^i) = w \}$.

\mypar{Feature Computation and Clustering}
For each weekday $w$ with arrivals (i.e., $\mathcal{T}_w^j \neq \emptyset$), we compute a feature vector $\mathbf{f}_w^j$ characterizing its arrival patterns, such as the average number of arrivals per day (Lines~\ref{line:compute_statistics}--\ref{line:standardize_features}). 
We then cluster\footnote{We use Ward's method~\cite{Ward_clustering}, but other clustering algorithms are also applicable.} the feature matrix $\mathbf{F}^j$ to group similar weekdays,
resulting in a mapping $\text{WTC}_j$ from weekdays to cluster labels (Line~\ref{line:ward_clustering}). Subsequently, the arrival times of the current global segment cluster $\mathbf{t}^i \in D_j$ are assigned to weekday clusters $W_k^j$ based on their weekdays' cluster labels (Line~\ref{line:weekday_cluster_definition}): $W_k^j = \{ \mathbf{t}^i \in D_j \mid \text{WTC}_j(\text{weekday}(\mathbf{t}^i)) = k \}$. Weekdays without arrivals (i.e., $\mathcal{T}_w^j = \emptyset$) are grouped into a separate cluster $W_{\text{NoData}}^j$ (Line~\ref{line:nodata_assignment}). The total number of weekday clusters $K_j$ equals the number of unique cluster labels in $\text{WTC}_j$, incremented by one if empty weekdays are present.
Finally, the set of weekday clusters over all global clusters $\mathcal{W} = \bigcup_{j=1}^{J} \{W_1^j, \dots, W_{K_j}^j\}$ represents the output of step 2, with each global cluster $D_j$ divided into disjoint subsets such that $D_j = \biguplus_{k=1}^{K_j} W_k^j$, with $W_k^j \in \mathcal{W}$. Since each subset $W_k^j$ is created by performing clustering within its global cluster $D_j$, the elements within a subset are more similar to each other than to those in any other subset, reflecting improved statistical coherence.



\subsection{Step 3: Intraday Binning}
\label{sec:intraday_binning}

This step addresses \emph{intraday dynamics} within each weekday cluster $W_k^j \in \mathcal{W}$ identified in~\autoref{sec:weekday_clustering}. To capture variations in arrival patterns throughout the day, we divide each day $\textbf{t}^i \in W_k^j$ into $L \in \mathbb{N}$ equally long time bins. Specifically, we partition the time domain of a day into $L$ consecutive intervals, where each bin spans an equal duration (e.g., 3 hours). For each day $\textbf{t}_{(j,k)}^i$ in $W_k^j$, we segregate the arrival times into these bins, resulting in subsets $\{\mathbf{t}_{(j, k,1)}^i,\dots, \mathbf{t}_{(j,k,L)}^i\}$, where: $\mathbf{t}_{(j,k,l)}^i$ represents all timestamps that occurred during the $l$-th bin on day $\textbf{t}_{(j,k)}^i$. Also, we obtain that each weekday cluster $W_k^j$ can now be represented via a disjoint union of the aforementioned bins: $W_k^j = \biguplus_{l=1}^{L} \mathbf{t}_{(j,k,l)}^i$.

\subsection{Step 4: Inter-arrival Learning via KDE}
\label{sec:fitting}
Having completed the \emph{divide} phase of our approach, we are able to fully partition the arrival dataset: $\mathcal{D} = \biguplus_{j=1}^J \biguplus_{k=1}^{K_j} \biguplus_{l=1}^{L} \mathbf{t}_{(j,k,l)}$. Concretely, each sequence of arrival times $\mathbf{t}_{(j,k,l)}$ corresponds to a global cluster $D_j$, a weekday cluster $W_k^j$, and an intraday bin $l$, and will be associated to a separate arrival time model. By dividing the dataset along these temporal dimensions, we expect to mitigate much of the underlying dynamic variability, thereby enabling an effective application of static modeling methods. However, in contrast to static methods that fit a parametric distribution to the inter-arrival times, we generalize this endeavor via Kernel Density Estimation (KDE); a flexible, non-parametric tool that learns the distribution directly from data. Briefly, the KDE for a probability density function $f$, given i.i.d. realizations $(X_i)_{i=1}^n$, is defined as $\hat{f}(x) = \frac{1}{n h} \sum_{i=1}^n K\left( \frac{x - X_i}{h} \right)$. Here, $h$ is termed \textit{bandwidth} and $K(\cdot)$ the \textit{kernel function}. The quality of the estimated density $\hat{f}$ depends on both $K$ and $h$ (cf.~\cite{kde}).

Modeling each subset's inter-arrival times with a separate KDE forms an \emph{ensemble of models} $\mathcal{E}$, which we use in the final step to generate new arrivals.

\subsection{Step 5: Arrival Generation}
\label{sec:data_gen}

The final step of our AT-KDE approach generates new arrival data for the simulation period. Algorithm~\ref{alg:data_generation} outlines this step, which consists of two main phases:

\mypar{Time Frame Initialization}
We first need to estimate which of the observed global segment clusters best represent the future simulation period. Thus, each to-be-simulated day $i\in \{1,\dots, \Tilde{N}\}$ is assigned to a global segment cluster label (Line~\ref{line:estim_segments}). For this, we propose a procedure following two rules: 1) If the sequence of observed global segment cluster labels $\mathcal{L}$ indicates a recurring pattern, we replicate it over the simulation period by determining the order and lengths of the segments and dividing the simulation period accordingly. For instance, if we have four global segments with cluster labels $\mathcal{L} = \langle1,2,1,2\rangle$, we can observe the recurring pattern that cluster $1$ is followed by $2$, which we then replicate accordingly. 2) If no recurring pattern is found (as in our motivation in \autoref{fig:corona}), we assume that the \emph{most recent} global segment cluster best represents the future. 

Then, we identify the earliest and latest times of the day when an arrival was observed in the training dataset $\mathcal{D}$ (Line~\ref{line:bounds}) to approximate the \emph{working hours} of the process. Subsequently, this range is divided into $L$ equally long time bins (Line~\ref{line:bin}), as also done in~\autoref{sec:intraday_binning}.

By default, our approach initiates the simulation of new arrivals at the start time of the test set and generates cases over the same time span. However, it also supports user-defined configurations, allowing for a custom simulation start time and a specified number of cases to be generated.

\begin{algorithm}[htbp]
\caption{GenerateArrivals}
\label{alg:data_generation}
\begin{algorithmic}[1]
    \REQUIRE Num. of days $\Tilde{N}$, Labels $\mathcal{L}$, Segments $\mathcal{S}$, Dataset $\mathcal{D}$, Bins $L$, KDE ensemble $\mathcal{E}$
    \ENSURE Generated arrival dataset $\mathcal{D}'$
    \STATE \COMMENTTRIANGLE{\textbf{Time Frame Initialization}}
    \STATE $\textit{estim\_segments\_per\_day} \gets \text{EstimateSegmentCluster}(\Tilde{N} , \mathcal{L}, \mathcal{S})$ \label{line:estim_segments}
    \STATE $\textit{lower\_time},\ \textit{upper\_time} \gets \text{DetermineBounds}(\mathcal{D})$ \label{line:bounds}
    \STATE $\textit{time\_bins} \gets \text{CreateTimeBins}(\textit{lower\_time},\ \textit{upper\_time}, L)$ \label{line:bin}
    \STATE \COMMENTTRIANGLE{\textbf{Arrival Data Generation}}
    \STATE $\mathcal{D}' \gets \langle \rangle$ \COMMENTTRIANGLE{Initialize sequence of arrival sequences}
    \FOR{$i \in \{1, \dots, \Tilde{N}\}$} \label{line:main_loop}
        \STATE $\textit{current\_date} \gets \text{GetDate}(i)$ \label{line:get_date}
        \STATE $\textit{weekday\_cluster} \gets \text{GetWeekdayCluster}(\textit{current\_date})$ \label{line:get_week_cluster}
        \STATE $\textit{estim\_segment} \gets \textit{estim\_segments\_per\_day}[i]$ \label{line:lookup}
        \STATE $\textit{seq} \gets \langle \rangle$ \COMMENTTRIANGLE{Initialize arrivals for one day}
        \FOR{each bin in $\textit{time\_bins}$} \label{line:bin_loop}
            \STATE $\textit{interarrivals} \gets \text{SampleInterarrivals}(\mathcal{E}, \textit{bin}, \textit{weekday\_cluster}, \textit{estim\_segment})$
            \STATE $\textit{arrivals} \gets \text{GenerateArrivals}(\textit{interarrivals},\ \textit{bin.start\_time},\ \textit{bin.end\_time})$
            \STATE $\textit{seq}.\text{append}(\textit{arrivals})$
        \ENDFOR
        \STATE $\mathcal{D}'.\text{append}(\textit{seq})$
    \ENDFOR
    \RETURN $\mathcal{D}'$
\end{algorithmic}
\end{algorithm}

\mypar{Arrival Data Generation}
We generate arrivals for each day $i\in \{1,\dots, \Tilde{N}\}$ in the simulation period (Line~\ref{line:main_loop}) by first determining the following parameters for our generation-method: the date corresponding to day $i$, beginning with the last date of the training data (Line~\ref{line:get_date}) per default, the corresponding weekday cluster (Line~\ref{line:get_week_cluster}), and the estimated global segment cluster (Line~\ref{line:lookup}). 

Given these parameters for day $i$, we iterate through each time bin of that day (Line~\ref{line:bin_loop}) and sample inter-arrivals from the corresponding KDE in $\mathcal{E}$, which are cumulatively summed to form the actual timestamps until the end of the bin is reached. Note that weekday clusters $W_{\textnormal{NoData}}^j$ based on absent arrivals yield no KDE model. Respective days are instead reflected via no arrivals.
The final output is the sequence of generated arrival timestamps $\mathcal{D}'$.

\section{Experiments and Results}
\label{sec:evaluation}
This section presents the experiments used to evaluate the performance of our AT-KDE approach for simulating arrival times. In the remainder, \autoref{sec:data_sum} 
describes the experimental setup, followed by the results in \autoref{sec:results}. Implementations and additional results can be found in our repository\footnote{\url{https://github.com/konradoezdemir/AT-KDE}}.



\subsection{Experimental Setup}
\label{sec:exp_setup}
\mypar{Evaluation Data}\label{sec:data_sum} Our evaluation is based on a diverse set of 20 event logs\footnote{Datasets available at \url{https://data.4tu.nl} or \url{https://zenodo.org/records/5734443}} spanning domains such as financial services, public administration, and healthcare. These logs vary widely in key characteristics (details available in our repository), including the number of arrivals, time spans, and arrival rates. For example, some processes exhibit pronounced global dynamics, such as abrupt increases in daily arrivals, while others remain relatively stable. Additionally, the logs feature a range of weekday and intraday patterns, from standard business hours (Monday to Friday, 9am–5pm) to continuous 24/7 arrivals, offering a comprehensive testing ground. For each event log, we derive the arrival dataset by selecting the earliest recorded timestamp for each case.

\mypar{Benchmark Approaches} We compare our approach against seven others, selected to ensure broad coverage of techniques from various domains:
\begin{itemize}[topsep=0pt]
    \item \emph{Mean}: A simple baseline using the mean inter-arrival time to model arrivals.
    \item \emph{Best Distribution}~\cite{Camargo_2020}: A static approach that selects the best fitting distribution out of a set of distributions to sample new arrival timestamps from.
    \item \emph{Prophet}~\cite{TaylorL17}: A dynamic time series forecasting method estimating hourly arrivals and distributing them uniformly within each hour to get timestamps.
    \item \emph{LSTM}: A recurrent neural network with \textit{LSTM} cell trained to predict the next inter-arrival time based on a prefix of arrivals and temporal features.
    \item \emph{Chronos}~\cite{Ansari2024}: A pretrained time series forecaster based on language models.
    \item \emph{XGBoost}: A gradient boosting model that predicts inter-arrival times using engineered temporal and contextual features in a regression task.
    \item\emph{NPP}~\cite{hawkes_in_queues}: Models arrivals as a Hawkes process with an exponential kernel, with the intensity function capturing baseline variations and self-excitation.
\end{itemize}

\noindent Note that we augment \emph{Mean, Best Distribution}, and \emph{NPP} with a probabilistic component to consider non-working days and first and last arrival timestamps within a day, aligning with their usage in simulation models~\cite{Camargo_2020,Kirchdorfer2024}. Initially, we considered ARIMA-based forecasting via, for example, \textit{auto-ARIMA}\footnote{\url{https://alkaline-ml.com/pmdarima}}. However, automatic parameter selection proved inconsistent across datasets, frequently yielding configurations that critically undermined the model's performance. Reliable ARIMA forecasts required manual tuning per dataset, contradicting our goal of a generalized approach; thus, we excluded ARIMA from our analysis. 

\mypar{Data Split} We perform a temporal hold-out split with the first 80\% of cases being in the training set and the last 20\% of cases in the test set.

\mypar{Hyperparameters} 
AT-KDE requires bandwidth optimization, which we initially estimate via Silverman's rule of thumb~\cite{Silverman86} using the first 80\% of the training data. This ground bandwidth is then multiplied with a factor $k$, where $k \in (0, 200]$, selecting the product-bandwidth that performs best on the remaining 20\% of the training set. Moreover, we opt for the Gaussian Kernel $K$ \cite{kde}. For clustering and segmentation, we adopt a sensitivity range $Z = [0.1,1]$, a temporal window $\omega=7$ days, a maximum $k_{\max}=6$ weekday clusters, and a bin size $L=3$. This configuration proved robust across diverse datasets, though practitioners may widen or narrow these settings (e.g., to accommodate different arrival sparsities). For both \emph{LSTM} and \emph{XGBoost}, grid-search-based hyperparameter tuning is performed. For \textit{Prophet}, we follow the tuning protocol of Camargo et al.~\cite{camargo_DSIM}. The parameters for \textit{NPP} are determined via maximum likelihood estimation. 


\mypar{Metrics}
To evaluate and compare the different arrival time prediction approaches, we use the Case Arrival Distribution Distance (CADD) metric proposed by Chapela-Campa et al. \cite{chapela2023}. This metric aggregates the test and simulated arrival timestamps into (separate) hourly counts and computes the Earth Mover's Distance between the resulting empirical distributions.


\subsection{Results}
\label{sec:results}

\begin{table}[t]
\centering
\setlength{\tabcolsep}{2.5pt} 
\caption{Average results of benchmarks measured by the square root of CADD. Missing entries based on convergence failure due to data size.}
\label{tab:results}
\begin{tabular}{lcccccccc}
\toprule
Event log & Mean & Best Dist. & Prophet & LSTM & Chronos & XGBoost & NPP & AT-KDE \\
\midrule
BPIC12 & 4.83  & 5.12  & 4.46  & 4.18 & 5.48 & 4.26 & 5.08  & \textbf{3.71}  \\
BPIC12CW & 6.61  & 6.84  & 4.47  & 6.27 & 8.33 & 8.69 & 6.79  & \textbf{3.21}  \\
BPIC12O & 14.75  & 14.55  & 13.61  & 15.00 & 11.72  &  12.73 & 14.59 & \textbf{11.56} \\
BPIC12W & 4.78 & 4.76  & 5.07  & 4.54 & 4.87  & \textbf{3.66} & 4.48 & 4.80  \\
BPIC13C & 66.22  & 66.65  & 13.99  & 27.93 & 14.14  & 13.39  & 68.70 & \textbf{13.25}  \\
BPIC17W & 15.63  & 15.47  & \textbf{9.65}  & 13.74 & 13.29  & 11.96 & 15.57 &  10.24 \\
BPIC19 & 300.66  & 295.53  & 17.83  & / & 22.54  &  21.38  & / & \textbf{17.58}  \\
BPIC20D & 10.33  & 11.55  & \textbf{5.95}  & 6.94 & 8.58  &  6.90 & 11.31  & 6.84  \\
BPIC20I & 28.02  & 29.09  & 17.41  & \textbf{15.01} & 19.93  & 17.37 & /  & 16.59  \\
BPIC20P & 25.06  & 24.47  & 15.08 & \textbf{10.91} & 13.58  & 12.58  & 25.14 & 13.82  \\
Env.permit & 33.48  & 31.99  & 17.97  & 19.62 & 18.19  &  14.98  & 31.76 & \textbf{13.74} \\
HelpDesk & 55.69  & 56.15  & 23.28  & 19.58 & 44.39  & 36.79  & 55.61 & \textbf{19.48} \\
Hospital & 22.59 & 22.45  & \textbf{15.89}  & 22.91 & 31.10  & 22.25 & 22.56 & 22.75  \\
Sepsis & 20.97 & 21.55  & 15.60  & 18.33 & 18.19  &  18.52  & 20.86 & \textbf{15.56} \\
P2P & 25.28  & 25.91 & 22.65  & 26.41 &  24.42  & 20.43  & 13.05 & \textbf{12.87} \\
CVS & 8.97  & 9.02  & 6.97  & 8.01 & 5.78  &  6.99  & 9.09 & \textbf{3.54} \\
Conf. 1000 & 12.54  & 12.25  & 6.26  & 7.16 &  6.05  &  9.68  & 12.16 & \textbf{5.20}  \\
Conf. 2000 & 18.23  & 17.97  & 9.86  & 11.33 &  10.15  &  8.99  & 18.30 &\textbf{6.49}  \\
ACR & 9.48 & 9.63  & 6.80  & \textbf{6.49 } & 8.63  & 7.34  & 8.78 & 7.00  \\
Production & 9.06  & 8.47  & 10.66  & 7.06 & 6.38  &  \textbf{4.15}  & 8.97 & 6.42 \\
\bottomrule

\end{tabular}
\end{table}

\mypar{Overall Results}
\autoref{tab:results} summarizes the experimental results for the 20 event logs over 10 runs. Our AT-KDE approach achieves the best performance in 12 logs---clearly outperforming the benchmarks---followed by \textit{Prophet}, \textit{LSTM}, and \textit{XGBoost} leading in the remaining 3/3/2 logs respectively. The static approaches \textit{Mean}, \textit{Best Distribution}, and \textit{NPP}, likely due to the rigidness of their parametric assumptions, lack the flexibility required to adapt effectively to temporal dynamics, failing to claim a single dataset. Notably, the recently proposed time series foundation framework \textit{Chronos} does not achieve a best performance on any of the datasets.
In all logs where AT-KDE is not the top performer, the gap to the leading approach is rather small. The only exception is \emph{Hospital}, where the gap to \emph{Prophet} is substantial. 
This can likely be attributed to a
sudden decrease in arrival counts during the log's test period. We believe that \emph{Prophet}'s focus on uncovering trends allowed it to anticipate this drop, whilst, for AT-KDE, the evidence hinting at this phenomenon appears too scarce. 
Overall, the results suggest that AT-KDE provides a robust approach for modeling arrival times in diverse and complex process datasets, not only outperforming static distributions but also more competitive dynamic approaches from other domains such as time-series and queueing theory.

\begin{figure}[htbp]
    \centering
    \includegraphics[trim = 0mm 0mm 0mm 0mm, clip, width=\columnwidth]{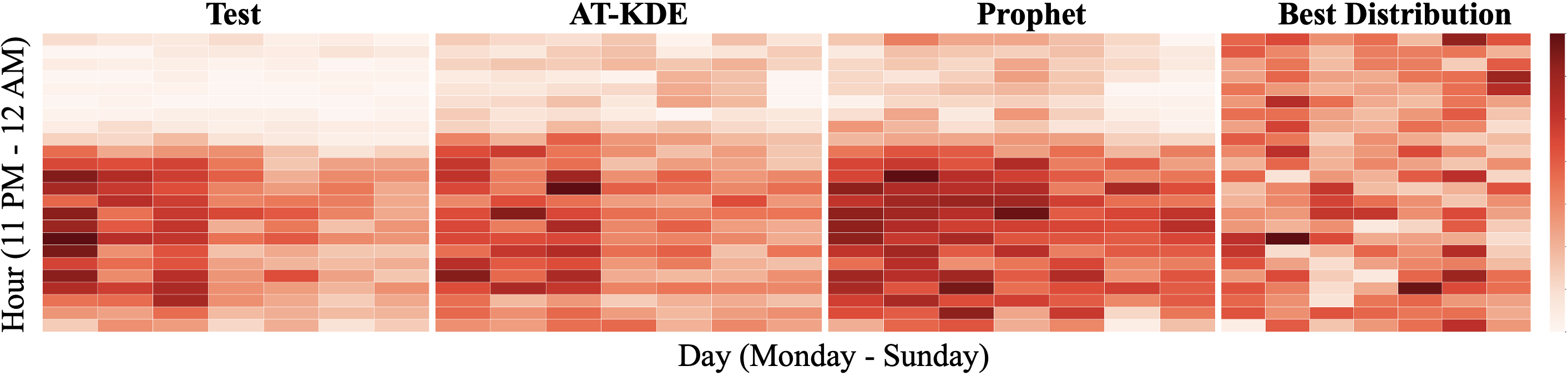}
    \caption{Distribution of arrivals of BPIC12 per hour of each day of the week.}
    \label{fig:day-heatmaps}
\end{figure}

\mypar{Outline of Weekday and Intraday Simulation Quality} 
To gain a better understanding of how the simulated arrival timestamps differ across approaches already used in BPS, we compare the hour-day distribution of arrivals in the BPIC12 log for 
AT-KDE, \emph{Prophet}, and \emph{Best Distribution} against the test data in \autoref{fig:day-heatmaps}. The process shows most arrivals between Monday and Wednesday, with only few at night. While \emph{Best Distribution} fails to capture this pattern, both AT-KDE and \emph{Prophet} are able to distinguish between day and night patterns. However, AT-KDE more accurately concentrates arrivals on the first three weekdays, whereas \emph{Prophet} overestimates arrivals from Thursday to Sunday.



\begin{figure}[t]
     \centering
     \begin{subfigure}[t]{0.49\textwidth}
         \centering
         \includegraphics[trim = 0mm 0mm 0mm 0mm, clip, width=\textwidth]{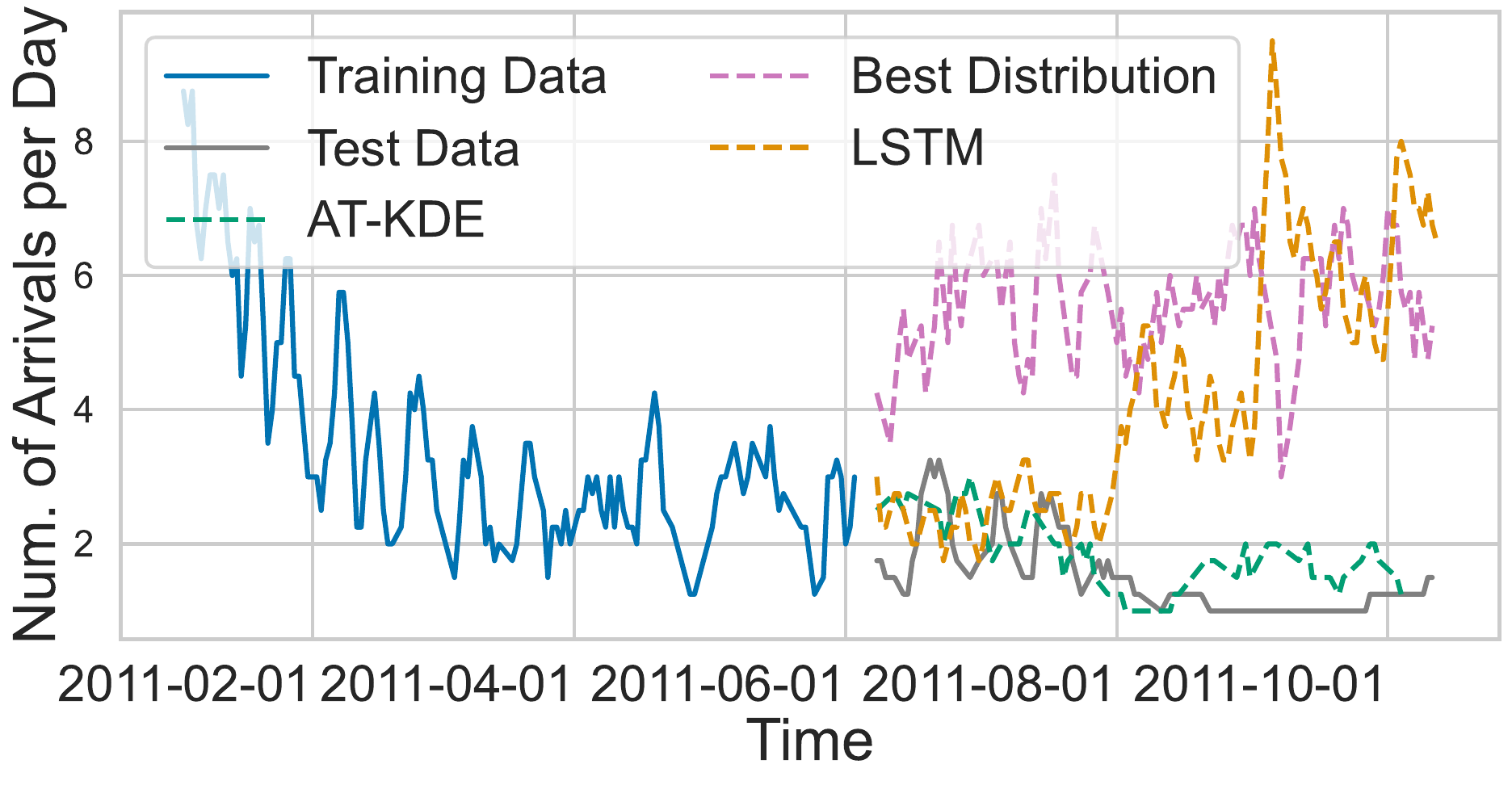}
         \caption{P2P}
         \label{fig:performance_insights_p2p}
     \end{subfigure}
     \hfill
     \begin{subfigure}[t]{0.49\textwidth}
         \centering
         \includegraphics[trim = 0mm 0mm 0mm 0mm, clip, width=\textwidth]{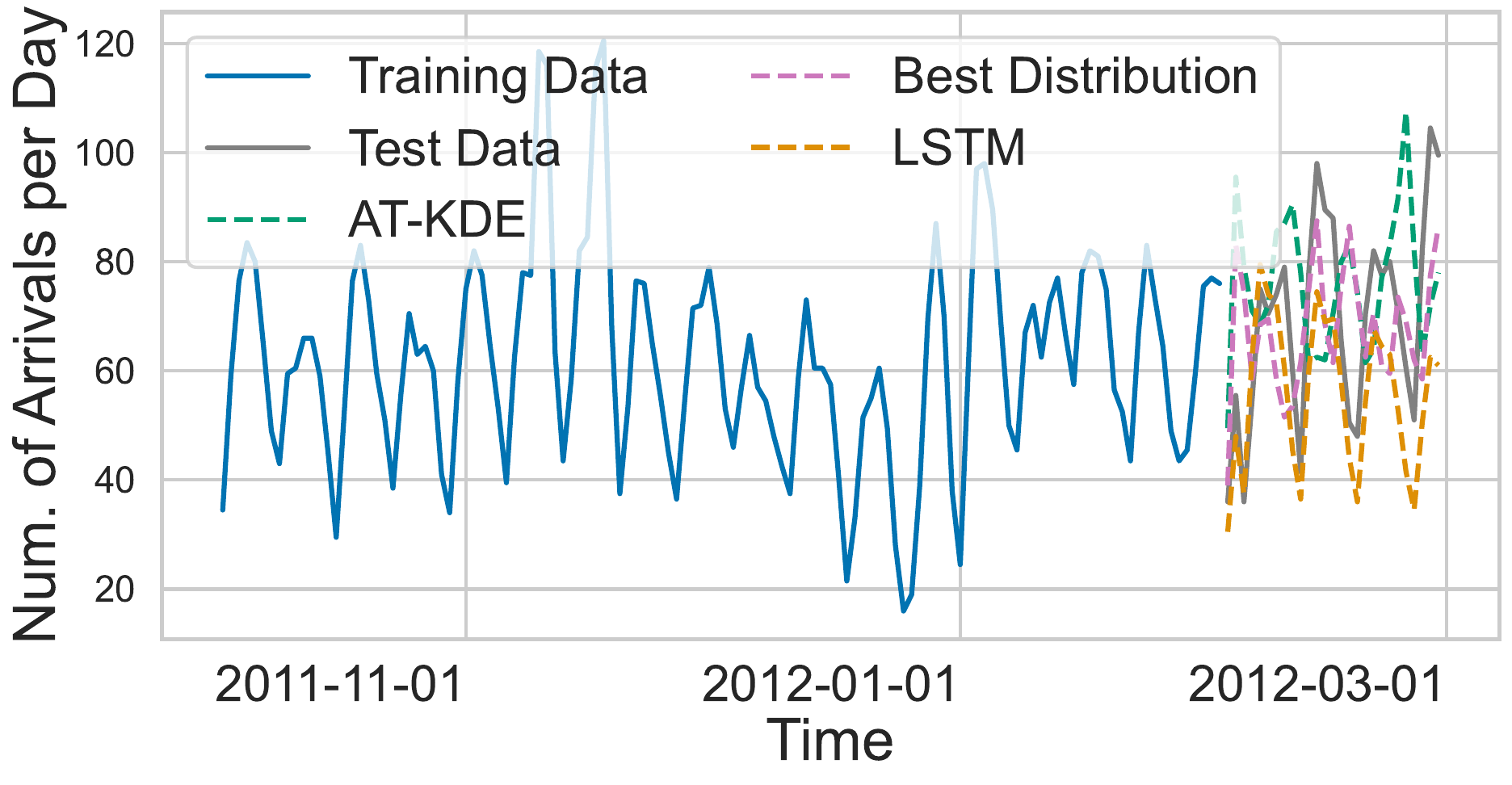}
         \caption{BPIC12W}
         \label{fig:performance_insights_bpic2w}
     \end{subfigure}
     \hfill
    \caption{Comparison of arrivals between AT-KDE, \textit{LSTM} and \textit{Best Distribution}.}
    \label{fig:performance_insights}
\end{figure}

\mypar{Post-hoc Analysis}
To examine in which scenarios AT-KDE outperforms existing approaches and when differences are negligible, we showcase two event logs in detail. For P2P, AT-KDE significantly outperforms the benchmarks. As shown in~\autoref{fig:performance_insights_p2p}, P2P is subject to a strong decrease in daily arrival counts during the first half of the training period. The static approach \textit{Best Distribution} ignores this drift and takes all training data into account, leading to a substantial overestimate of daily arrivals compared to the actual test data.
Notably, even a dynamic approach like the \emph{LSTM} fails to produce realistic arrival timestamps, as it appears biased by the high arrival volumes at the start of the training period. In contrast, AT-KDE correctly identifies this drop and explicitly considers it by leveraging only post-drift data for the simulation, yielding significantly more accurate arrival times.
Conversely, BPIC12W (\autoref{fig:performance_insights_bpic2w}) shows no drift behavior on the global scale with arrival counts exhibiting a rather constant mean and variance throughout time. Therefore, a static approach such as \textit{Best Distribution} can achieve comparable performance as \emph{LSTM} or AT-KDE in this scenario.


\mypar{Execution Time} 
All experiments were run on an Apple M3 Pro (12-core CPU, 18GB RAM). As suggested by \autoref{tab:runtime}, \emph{Mean} and \emph{Best Distribution} achieve near-instantaneous times. Although AT-KDE’s training takes slightly longer (13s), its simulation is competitively fast (0.03s). In contrast, all other benchmarks are notably slower in both phases, highlighting AT-KDE’s overall efficiency.\footnote{The full overview including standard deviations can be found in our repository.}

\begin{table}[htbp]
\centering
\setlength\tabcolsep{3pt} 
\caption{Mean execution times in seconds across all 20 logs.}
\label{tab:runtime}
\begin{tabular}{l|cccccccc}
\toprule
 & \textbf{Mean} & 
 \textbf{BD} & 
 \textbf{Prophet} & 
 \textbf{LSTM} & 
 \textbf{Chronos} & 
 \textbf{XGB} & 
 \textbf{NPP} & 
 \textbf{AT-KDE} \\
\midrule

Training 
& 0.02 
& 1.76 
& 745.19 
& 25.45 
&  102.37 
& 77.67 
& 270.09 
& 13.16 
\\

Simulation 
& 0.03 
& 0.13 
& 174.73 
& 337.55 
& 0.58 
& 26.66 
& 0.31 
& 0.03 
\\
\bottomrule
\end{tabular}
\end{table}


\section{Conclusion}
\label{sec:conclusion}
In this work, we introduced \emph{Auto Time Kernel Density Estimation} (AT-KDE), an approach for learning a case-arrival model from data and generating new arrival times for BPS. AT-KDE employs a divide-and-conquer strategy combined with a kernel density estimator to effectively capture global, weekday, and intraday arrival dynamics. By partitioning the input arrival dataset into subsets and modeling each with a dedicated KDE, our approach can generate new arrival timestamps that reflect the dynamic nature of organizational processes, enabling more accurate simulation outcomes. Extensive evaluations highlight the critical role of incorporating these temporal dynamics, with AT-KDE delivering superior accuracy and robustness compared to a variety of existing approaches, all while maintaining sensible runtime efficiency.

\mypar{Limitations}
Despite its strong empirical performance, AT-KDE is still confronted with limitations. First, its effectiveness relies on careful calibration tailored to each process to ensure optimal performance. Also, while we demonstrated that accurate arrival estimates improve overall simulation quality, the evaluation is restricted to arrival estimation due to missing event start timestamps in the process logs and the scope of this work. Moreover, in the absence of clear training‐data patterns, AT-KDE falls back on recent behavior to predict future arrivals---a heuristic that, while usually adequate, can miss subtler trends captured by more complex deep-learning models.


\mypar{Future Work}
While initially developed for modeling arrival times, our divide-and-conquer approach may be extended to capture further process dynamics, such as evolving activity durations or delays. Furthermore, its underlying principles are broadly applicable beyond BPS, offering value in any domain where time-dependent behavior plays a critical role in arrival modeling. Moreover, while we and related works give empirical evidence of the positive impact of accurate arrival modeling on overall simulation quality, the research community would benefit from a comprehensive analysis of this assumed correlation across a broad range of real-life event logs.

%
%
\bibliographystyle{splncs04}
\bibliography{bibliography}
%


\end{document}